%% file: main.tex
\documentclass{article}

\usepackage{microtype}
\usepackage{graphicx}
\usepackage{booktabs}

\usepackage{hyperref}

\usepackage[accepted]{icml2023}

\usepackage{amsmath}
\usepackage{amssymb}
\usepackage{mathtools}
\usepackage{amsthm}

\usepackage{microtype}
\usepackage{array}
\usepackage{multirow}
\usepackage{tipa}
\usepackage{url}
\usepackage{caption}
\usepackage{bm}
\usepackage{subcaption}
\usepackage{hyperref}
\usepackage{cite}
\usepackage{float}
\usepackage{soul}
\usepackage{anyfontsize}

\usepackage[capitalize,noabbrev]{cleveref}

\def\discrete{\textsc{Discrete}}
\def\acousticee{\textsc{Acoustic*~E2E}}
\def\acousticees{\textsc{Acoustic~E2E}}
\def\acousticpr{\textsc{Acoustic*~+~PhoneRec}}
\def\acousticprs{\textsc{Acoustic~+~PhoneRec}}

\theoremstyle{plain}

\theoremstyle{definition}

\theoremstyle{remark}

\icmltitlerunning{Learning to Speak and Hear Through Multi-Agent Communication}

\begin{document}

\twocolumn[
\icmltitle{Towards Learning to Speak and Hear Through Multi-Agent Communication over a Continuous Acoustic Channel}

\icmlsetsymbol{intern}{*}

\begin{icmlauthorlist}
\icmlauthor{Kevin Eloff}{su,insta,intern}
\icmlauthor{Okko Räsänen}{fin}
\icmlauthor{Herman A. Engelbrecht}{su}
\icmlauthor{Arnu Pretorius}{insta}
\icmlauthor{Herman Kamper}{su}
\end{icmlauthorlist}

\icmlaffiliation{insta}{InstaDeep, Cape Town, South Africa}
\icmlaffiliation{su}{E\&E Engineering, Stellenbosch University, South Africa}
\icmlaffiliation{fin}{Unit of Computing Sciences, Tampere University, Finland. *Work done during an internship at InstaDeep}

\icmlcorrespondingauthor{Kevin Eloff}{kevin.eloff@gmail.com}

\icmlkeywords{Language emergence, spoken communication, multi-agent reinforcement learning, language acquisition}

\vskip 0.3in
]

\printAffiliationsAndNotice{}  

\begin{abstract}
Multi-agent reinforcement learning has been used as an effective means to study emergent communication between agents, yet little focus has been given to continuous acoustic communication.
This would be more akin to human language acquisition; human infants acquire language in large part through continuous signalling with their caregivers.
We therefore ask: Are we able to observe emergent language between agents with a continuous communication channel?
Our goal is to provide a platform to begin bridging the gap between human and agent communication, allowing us to analyse continuous signals, how they emerge, their characteristics, and how they relate to human language acquisition.
We propose a messaging environment where a Speaker agent needs to convey a set of attributes to a Listener over a noisy acoustic channel.
Using DQN to train our agents, we show that: (1) unlike the discrete case, the acoustic Speaker learns redundancy to improve Listener coherency, (2) the acoustic Speaker develops more compositional communication protocols which implicitly compensates for transmission errors over a noisy channel, and (3) DQN has significant performance gains and increased compositionality when compared to previous methods optimised using REINFORCE.
\end{abstract}

\input{s1_introduction}
\input{s2_environment}
\input{s3_approach}
\input{s4_experiments}
\input{s5_discussion}

\bibliography{main}
\bibliographystyle{icml2023}

\end{document}

%% file: s1_introduction.tex
\section{INTRODUCTION}
\label{sec:introduction}

Reinforcement learning (RL) is increasingly being used as a tool to study language emergence
\citep{igor-pieter-2017-emergence, lazaridou-etal-2018, eccles-etal-2019-biases, chaabouni-etal-2020-compositionality, lazaridou-baroni-2020-emergent, auersperger-pecina-2022-defending}.
By allowing multiple agents to communicate with each other while solving a common task, a communication protocol needs to be established.
The resulting protocol can be studied to see if it adheres to properties of human language, such as compositionality and redundancy~\citep{kirby-2001-spontaneous, geffen-lan-etal-2020-spontaneous,andreas-2020-good,resnick-etal-2019}.
The tasks and environments themselves can also be studied, to see what types of constraints are necessary for human-like language to emerge~\citep{steels-1997-synthetic}. Referential games are often used for this purpose~\citep{kajic-etal-2020-learning, havrylov-titov-2017-emergence, yuan-etal-2020-emergence}.
While these studies open up the possibility of using computational models to investigate how language emerged and how language is acquired through interaction with an environment and other agents, most RL studies consider communication using \textit{discrete} symbols.

Spoken language instead operates and emerged over a \textit{continuous} acoustic channel.
Human infants acquire their native language by being exposed to speech audio in their environment~\citep{kuhl-2004-early}.
By interacting and communicating with their caregivers using continuous signals, 
{infants can observe the consequences of their communicative attempts (e.g.\ through parental responses) that may guide the process of language acquisition (see e.g.\ \citep{howard-etal-2014-learning} for a more in-depth discussion).} 
Continuous signalling is challenging since an agent needs to be able to deal    with different acoustic settings and noise introduced through the channel.
These intricacies are lost when agents communicate directly with purely discrete symbols.
This raises the question: Are we able to observe emergent language between agents with a {continuous} communication channel, trained using RL?
This paper is our first step towards answering this larger research question.

Earlier work has considered models of human language acquisition using continuous signalling between a simulated infant and caregiver~\citep{oudeyer-2005-the, steels-belpaeme-2005-coordinating}.
However, these models often rely on heuristic approaches and older neural modelling techniques, making them difficult to extend. For example, it is not easy to directly incorporate other environmental rewards or interactions between multiple agents.
More recent RL approaches would make this possible, but as noted, have mainly focused on discrete communication.
Our work tries to bridge the disconnect between recent contributions in multi-agent reinforcement learning (MARL) and earlier literature in language acquisition and modelling~\citep{moulinfrier-hal-2021-multiagent}.

\begin{figure}[!t]
    \centering
    \hspace*{-0.6cm}\includegraphics[width=1.2\linewidth]{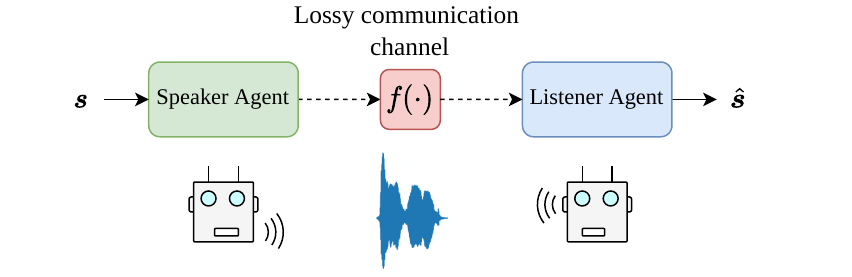}
    \caption{Environment setup where a Speaker communicates to a Listener over a lossy acoustic communication channel $f$.}
    \vspace{-0.5cm}
    \label{fig:simple}
\end{figure}

\begin{figure*}[!t]
    \centering
    \hspace*{0.4cm}\includegraphics[width=0.9\textwidth]{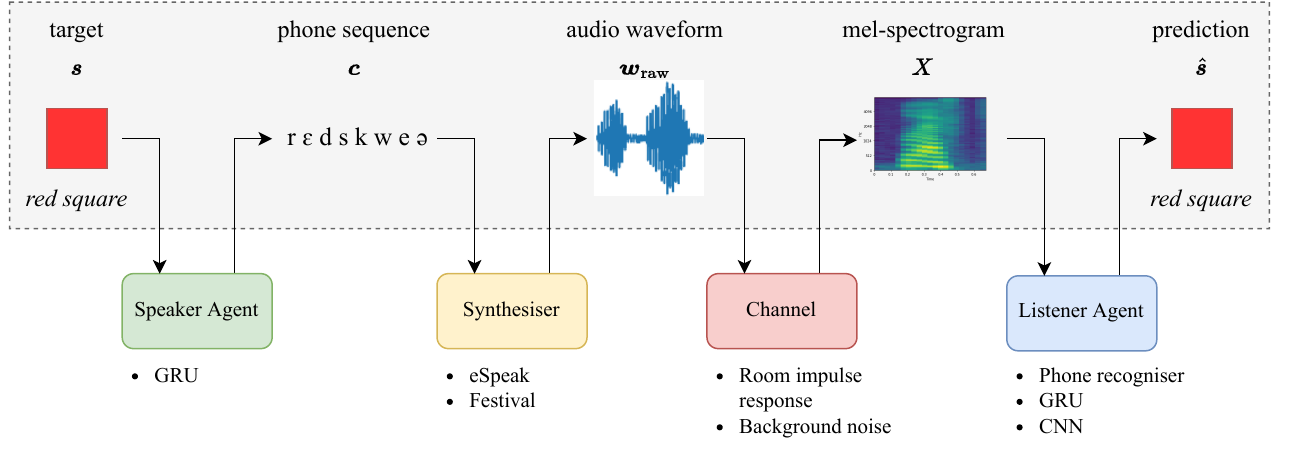}\vspace{-0.2cm}
    \caption{
        Example interaction of each component and the environment in a single round. 
        The Speaker observes the target \textit{red square} {(the attribute values that need to be communicated)} and generates a phone sequence.
        This phone sequence is synthesised, passed through a realistic acoustic channel, and finally received by the Listener.
        The Listener then correctly predicts \textit{red square}.
    }
\label{fig:envstep}
\end{figure*}

One recent exception which does use continuous signalling within a modern RL framework is \citep{gao-etal-2020-spoken} and the follow-up work~\citep{komatsu-2022-automatic}.
In these studies, a Student agent is exposed to a large collection of unlabelled speech audio, from which it builds up a dictionary of possible spoken words.
The Student can then select segmented words from its dictionary to play back to a Teacher, which uses a trained automatic speech recognition (ASR) model to classify the words and execute a movement command in a discrete
environment.
The Student is then rewarded for moving towards a goal position.
We also propose a Student-Teacher setup,
but importantly, our agents can generate their own unique audio waveforms rather than just segmenting and repeating words exactly from past observations.

Concretely, we propose the environment illustrated in Figure~\ref{fig:simple}, which is an extension of a referential signalling game used in several previous  studies~\citep{david-1969-convention,lazaridou-etal-2018,chaabouni-etal-2020-compositionality,rita-etal-2020-lazimpa}.
Here $\boldsymbol s$ represents a set of attribute values the {Speaker} must communicate to a {Listener} agent. 
Taking these attributes as input, the Speaker produces a waveform as output, which passes over a lossy
acoustic channel.
The Listener ``hears'' the utterance from the {Speaker}. 
Taking the waveform as input, the Listener produces output $\hat{\boldsymbol s}$.
This output is the Listener's interpretation of the concept that the Speaker agent tried to communicate.
The agents must develop a
common communication protocol such that $\boldsymbol s =\hat{\boldsymbol s}$. 
This process encapsulates one of the core goals of human language: conveying meaning through communication \citep{dor-2014-instruction}.

{To train the agents, we use a deep Q-network (DQN)~\citep{mnih-etal-2013-playing,mnih-etal-2015-human}.
{This differs from previous studies on language emergence~\citep{lazaridou-etal-2018, chaabouni-etal-2020-compositionality, rita-etal-2020-lazimpa, auersperger-pecina-2022-defending}}
that used REINFORCE~\citep{williams-1992-reinforce}.
REINFORCE is a policy gradient method known to have relatively high variance \citep{weaver-tao-2001-optimal}, while DQN is a value-based method which tends to have higher stability.
We find DQN to converge faster and more consistently, with
{the resulting communication protocols showing more}
compositionality and {better} generalisation than REINFORCE for the same models.}

Our bigger goal is to explore the question of whether and how language emerges when using RL to train agents that communicate via continuous acoustic signals.
Our proposed environment and training methodology serves as a means to perform such an exploration, and the goal of this paper is to showcase some capabilities of the platform.
{As a concrete example of the types of research questions we can answer with our environment,}
we consider how discrete and acoustic emergent language differs when agents communicate with sequences of different duration.
We show that when longer sequences are allowed, our acoustic {Speaker} learns a redundant communication protocol, using fewer unique phones per utterance with more repetition of bigrams and trigrams.
This redundancy improves the coherency of the {Listener} agent.
{In contrast, agents} trained to communicate with purely discrete symbols fail to learn such redundancy, resulting in poor communication when these agents are used in very noisy environments.
The acoustic agents also tend to develop more compositional communication protocols that adapt to transmission errors over the noisy acoustic channel.

%% file: s2_environment.tex
\section{ENVIRONMENT}
\label{sec:environment}

We base our environment on the referential signaling game from \citep{chaabouni-etal-2020-compositionality} and \citep{rita-etal-2020-lazimpa}---which itself is based on \citep{david-1969-convention} and \citep{lazaridou-etal-2018}---where a sender must convey a message to a receiver.
In our case, communication takes place between a Speaker and a Listener over a continuous acoustic channel, instead of sending discrete symbols directly (Figure~\ref{fig:simple}).
Formally, in each episode, the environment generates a set of attributes $\boldsymbol s = (s_1, s_2, \ldots, s_N$), 
defined as $N$ one-hot encoded vectors each representing one of $M$ attribute values, i.e.\ $s_i \in \{ 0, 1 \}^M$.
The Speaker receives $\boldsymbol s$ and generates a sequence of phones $\boldsymbol c = (c_1, c_2, \ldots, c_L)$,
each $c_t \in \mathcal{P}$ representing a phone from a predefined phonetic 
alphabet $\mathcal{P}$.
The phone sequence is then converted 
into a waveform $\boldsymbol w_\text{raw}$, an audio signal sampled at $16~\text{kHz}$.
For generating the Speaker's phone sequence,
we use a trained text-to-speech model~\citep{black-lenzo-2000-building,duddington-espeak}.
A channel noise function is then applied to the generated waveform, and the result $\boldsymbol w_\text{in} = f(\boldsymbol w_\text{raw})$ is presented as input to the Listener.
The Listener converts the input waveform to a mel-scale spectrogram: 
a sequence of vectors over time representing the frequency content of an audio signal
scaled to mimic human frequency perception~\citep{davis-mermelstein-1980-melspec}.
Taking the mel-spectrogram sequence $X = (\bm x_1, \bm x_2, \ldots, \bm x_T)$ of $T$ acoustic frames as input, the
Listener agent
outputs a vector $\hat{\boldsymbol s}$ representing its 
prediction of the attribute values.
The agents are both rewarded if the predicted vector of attributes is equal to the target vector of attributes $\boldsymbol s = \hat{\boldsymbol s}$.

To make the environment a bit more concrete, we present a brief example in
Figure~\ref{fig:envstep}. 
Consider a case where we transmit two attributes ($N=2$){:  one colour and one shape. Each attribute can take on one of}
three values ($M=3$):
$s_1 \in \{\text{\textit{red}}, \text{\textit{green}}, \text{\textit{blue}}\}$ and $s_2 \in \{\text{\textit{circle}}, \text{\textit{square}}, \text{\textit{triangle}}\}${.}
The concatenated state representation for \textit{red square} would be
$\boldsymbol s = \left[1, 0, 0, 0, 1, 0 \right]^\top$.
A possible phone sequence generated by the Speaker
{could}
be
$\boldsymbol {c} = (\text{\textipa{r}}, \text{\textipa{E}}, \text{\textipa{d}}, \text{\textipa{s}}, \text{\textipa{k}}, \text{\textipa{w}}, \text{\textipa{e}}, \text{\textipa{@}}, \text{\textless/s\textgreater})$.
This would be synthesised, passed through the channel, and then be interpreted by the Listener agent.
If the Listener's prediction is $\hat{ \boldsymbol s}  = \left[1, 0, 0, 0, 1, 0 \right]^\top$, then it correctly interpreted {the message as conveying the attributes} \textit{red square}.
The environment would then reward {both agents}. 

The environment reward is calculated as $R = \frac{1}{N}\sum_{i=1}^N r_i$, where $r_i$ is the per-attribute reward:
\begin{equation*}
r_i = \begin{cases} 
      1 & \text{if\ }  s_i =  \hat{  s}_i \\
      0 & \text{otherwise}
\end{cases}
\label{eq:r_component}
\end{equation*}

Since we use an existing text-to-speech system for the Speaker's output, the Speaker's task has in effect been converted to a discrete communication problem.
However, the combination of both agents and their environment is {still} a continuous communication task:
the noisy channel is based on real acoustic environments and the Listener takes in a noisy acoustic signal {(see Section~\ref{sec:acoustic_channel} for full details)}. 
What we have done here is to equip the Speaker with articulatory capabilities so that these do not need to be learned by the model.
Some studies consider how articulation can be learned~\citep{howard-etal-2014-learning, asada-2016-modeling, rasilo-rasanen-2017-online}, but none of these do so in an RL environment, rather using a form of imitation learning. 
In Section~\ref{sec:discussion}, we discuss how future work could consider learning the articulation process itself within our environment, and the challenges involved in doing so.

%% file: s3_approach.tex
\section{LEARNING TO SPEAK AND HEAR USING RL}

In this section, we first discuss the model implementations that allow us to speak and hear.
We then cover the communication channel and how we optimise our agents.
To train our agents, we {compare}
REINFORCE \citep{williams-1992-reinforce} (used in previous language emergence studies) to DQN \citep{mnih-etal-2013-playing}.
Finally, we cover other important implementation details, such as the model parameters, environmental setup, and discrete baseline system.

\subsection{Speaker model}
\label{sec:speaker_model}

The Speaker agent is tasked with generating a sequence of phones $\boldsymbol c  = (c_1, c_2, \ldots, c_L)$ describing a set of attributes.
The set of target attributes is represented by the one-hot input state $\boldsymbol s$.
We use gated recurrent unit (GRU)~\citep{cho-etal-2014-properties} based sequence generation as the core of the Speaker agent.
{The Speaker's GRU}
generates a sequence of logits.
{At each output-step (from 1 to $L$), a softmax over the logits defines a distribution over the set of phone symbols $\mathcal{P}$.}
During training, the symbols are sampled based on this distribution to produce $\boldsymbol c$.
At test time, $\boldsymbol c$ is greedily selected from the logits.
The input state $\boldsymbol s$ is embedded through a linear layer as the initial hidden state of the GRU.
The Speaker is allowed to generate phone sequences of arbitrary length, up to a maximum of $L$.

\subsection{Listener model}
\label{sec:listener_model}

Given an input mel-spectrogram $X$, the Listener generates a set of predicted attributes $\boldsymbol{\hat{s}}$.
The model, roughly based on \citep{amodei-etal-2016-DS2}, first applies a set of convolutional layers over the input mel-spectrogram, keeping the size of the time-axis consistent throughout. 
The convolution outputs {are} then flattened over the filters and feature axes, resulting in a single vector per time step.
Each vector is processed through a GRU, with a linear layer applied to the final hidden state to produce logits over attributes.
An argmax of these logits gives us a greedy prediction for $\hat{\boldsymbol s}$. 
We call this the ``{end-to-end} acoustic Listener'', as there are no intermediary steps going from the mel-spectrogram input to $\hat{\boldsymbol s}$.

As an alternative to the above approach, we simplify the task of the {Listener}:
we first process $X$ through a pre-trained static phone recogniser.
The model of the phone recogniser is also based on~\citep{amodei-etal-2016-DS2}.
This model is trained to perform connectionist temporal classification \citep{graves-2006-ctc} on arbitrary phone sequences in a noiseless environment. 
The phone recogniser {has} a phone error rate (PER) of $6.42\%$.
These phones are then consumed by the {Listener} agent, embedded with a linear layer, and then processed by a GRU.
The final hidden state of the GRU is passed through a linear layer and an argmax to arrive at our final greedy prediction for $\hat{\boldsymbol s}$.
{In our experiments, we denote this approach with a ``\textsc{+~PhoneRec}'' label.}
With this agent, communication is still different to the purely discrete case, since both the Speaker and Listener needs to develop a protocol that can compensate for information loss over the continuous communication channel (described next).

\subsection{Realistic communication channel}
\label{sec:acoustic_channel}

The lossy communication channel $f$ consists of two core components: superimposed background noise and convolution with a room's impulse response.
We first add background noise directly to the raw audio signal.
The background noise is randomly sampled from the Clotho audio captioning dataset \citep{drossos-2020-clotho}.
We then convolve the resulting waveform with a room's impulse response from the Aachen Impulse Response (AIR) Database \citep{jeub-2009-air}.
We use {five} 
rooms in total. In training, we use the \textit{booth}, \textit{lecture} and \textit{office} rooms, while for testing, we use the
\textit{meeting} and \textit{stairway} rooms separately as {unseen} evaluation rooms.
The full channel function $f$ is shown in Equation~\ref{eq:f}, where $\boldsymbol n \in \mathcal N$ is sampled from the Clotho dataset $\mathcal N$, $\boldsymbol h  \in \mathcal H$ {is} sampled from the AIR database $\mathcal H${, and $\ast$ denotes convolution.\footnote{{In this equation we are overloading the notation, with all vectors 
{representing}
discrete-time sequences.}}}

\vspace{-0.25cm}
\begin{equation}
f(\boldsymbol w_\text{raw}) = (\boldsymbol w_\text{raw} + \boldsymbol n)\ast \boldsymbol h
\label{eq:f}
\end{equation}

\subsection{Optimisation}
\label{sec:optim}

Most previous emergent language studies use REINFORCE \citep{williams-1992-reinforce}, a policy-gradient algorithm, for Speaker optimisation. 
The algorithm is used to optimise the policy: a model that generates a distribution over actions given an input state such that the actions taken maximise the expected reward.
REINFORCE is known to have high variance and often struggles to consistently converge \citep{weaver-tao-2001-optimal}.
Therefore, we compare REINFORCE to DQN~\citep{mnih-etal-2013-playing}, a more modern algorithm that first reached human-level performance on the now popular Atari benchmark \citep{mnih-etal-2015-human}.

We can also use DQN for Listener optimisation, but found it to be slightly less stable than
the approach from previous studies where the cross-entropy between $\boldsymbol s$ and the Listener's prediction of $\hat{\boldsymbol s}$ is optimised.
Therefore, to be consistent with prior work, we also use cross-entropy to optimise the Listener.
{In all experiments, the Speaker and Listener are trained simultaneously.}

We implement our acoustic environment and DQN in the EGG toolkit \citep{kharitonov-etal-2019-egg}.

\subsection{Implementation}
\label{sec:implementations}

For our Speaker agent, we use eSpeak~\citep{duddington-espeak} as our speech synthesiser. 
eSpeak is a parametric text-to-speech
software package that uses formant synthesis to generate audio from phone sequences. 
We also experimented with using Festival~\citep{black-lenzo-2000-building}, but
instead favoured eSpeak due to its {fast inference,} simpler phone scheme, and multi-language support.
In our experiments, the use of an existing speech synthesiser allows us to focus on the emergence of phonotactic and lexical structure in multiagent interaction without having to simultaneously focus on articulatory learning (learning to produce individual speech sounds).
In future work we may look to relax some of these assumptions: see Section~\ref{sec:discussion} for {a complete} discussion.

{In each communication round, the Speaker is allowed to generate up to $L$ tokens.}
All GRUs {are} {single layered} with a hidden layer size of 256. 
The default phone set used is $\mathcal{P} = \{\text{\textipa{a}}, \text{\textipa{e}}, \text{\textipa{i}}, \text{\textipa{o}}, \text{\textipa{u}}\}$
These are a subset of the available eSpeak phones, denoted here using the international phonetic alphabet (IPA).
All experiments are performed with a fixed input size of $N=4$ and $M=5$ {attributes}, giving 
a total input size $|S| = M^N$ of 625 attribute combinations.
{All models are trained until convergence, which always occurs before 50 training epochs.
All other setup settings match 
the
default configuration of the \texttt{compo\_vs\_generalization}~\citep{chaabouni-etal-2020-compositionality} environment in EGG~\citep{kharitonov-etal-2019-egg}.}

As a baseline solution, we train the Speaker and Listener models of \citep{chaabouni-etal-2020-compositionality} to solve the discrete communication task.
During training, the discrete units generated by the Speaker are directly consumed by the Listener.
During evaluation, we follow the same procedure as our acoustic implementation (Section~\ref{sec:environment}).
We first map the discrete units of the discrete Speaker to the same phone set $\mathcal P$ used by our acoustic Speaker, thereafter utilising an identical synthesis and channel noise setup.
A pre-trained phone recogniser (described in Section~\ref{sec:listener_model}) is then used to interpret the waveform as discrete units which are fed to the discrete Listener.
This baseline thus represents the setting where discrete-only agents are trained and then employed in an acoustic environment using fixed symbol-to-waveform and waveform-to-symbol models.

We compare five main system variants in our experiments:
\begin{enumerate}
    \item {\discrete{}}:
    Our baseline follows the discrete implementation described above,
    where we first train the agent to solve the discrete communication task.
    The agent is then evaluated in the acoustic communication channel of Section~\ref{sec:acoustic_channel}.
    To do this, the discrete units are mapped to phones, which are synthesised and passed over the channel. A pre-trained phone recogniser (Section~\ref{sec:listener_model}) then converts back to discrete units for the Listener.
    \item {\acousticees{}}:
    This approach combines our acoustic Spea-ker model (Section~\ref{sec:speaker_model}) with the end-to-end (\textsc{E2E}) Listener model (Section~\ref{sec:listener_model}).
    As a reminder, the end-to-end Listener converts directly from a mel-spectrogram to the predicted attributes~$\hat{\boldsymbol s}$.
    \item {\acousticee{}}:
    We consider a variant of the above model 
    where the agents are pre-trained; in this case the Speaker agent is initialised with the weights of the \discrete{} Speaker. We denote system variants that use pre-training with a `*'.
    \item {\acousticprs{}}:
    Our fourth approach combines the same acoustic Speaker model with the phone recogniser (\textsc{PhoneRec}) Listener (Section~\ref{sec:listener_model}).
    The phone recogniser used here is identical to that of the \discrete{} approach.
    We use this variant as a direct comparison to the \discrete{} model, as neither are able to update the weights of the static phone recogniser.
    \item {\acousticpr{}}:
    The final approach uses pre-training with \acousticpr{}.
    In this approach, both the Speaker and Listener are initialised with the weights of the \discrete{} agents.
\end{enumerate}

\begin{table*}[!t]
    \caption{{Per-attribute accuracy and topographic similarity of various models in both the training and evaluation environments. \textsc{Acoustic*} uses the discrete baseline for pre-training. Each model was trained five times. We observe a maximum standard deviation of 0.02 over all seeds.}}
\vspace{-.2cm}
    \centering
    \small
    \begin{tabular}{c l c c c c c c}
        \toprule
        && \multicolumn{3}{c}{REINFORCE} & \multicolumn{3}{c}{DQN} \\
        \cmidrule(lr){3-5} \cmidrule(lr){6-8}
        &\textit{Model} & \textit{train rooms} & \textit{eval.\ rooms} & \textit{topsim}  & \textit{train rooms} & \textit{eval.\ rooms} & \textit{topsim} \\
        \midrule
        1 & \discrete{} & 0.621 & 0.612 & 0.387 & 0.649 & 0.649 & 0.691 \\
        2 & \acousticees{} & 0.611 & 0.566 & 0.275 & 0.956 & 0.789 & 0.519 \\
        3 & \acousticee{} & {0.973} & {0.950} & 0.373 & \textbf{0.986} & 0\textbf{.958} & 0.707 \\
        4 & \acousticprs{} & 0.539 & 0.535 & 0.358 & 0.682 & 0.674 & \textbf{0.747} \\
        5 & \acousticpr{} & 0.609 & 0.591 & 0.387 & 0.726 & 0.710 & 0.745 \\
        \bottomrule
    \end{tabular}
    \label{tab:models}
\end{table*}

%% file: s4_experiments.tex
\section{EXPERIMENTS}

\subsection{Different approaches in noisy environments}
\label{sec:approaches}

We start by comparing different system variants in noisy environments, with the goal of seeing how discrete-only training compares with training an acoustic agent.
As a reminder, the discrete agents are trained exclusively for symbolic communication, while the acoustic agents are trained in an environment with channel noise.

\textbf{Experimental setup.}
In these experiments, all agents use a fixed sequence length of $L=5$. 
The acoustic models are trained with the full lossy communication channel described in Section~\ref{sec:acoustic_channel}, including background noise samples {from Clotho} {that are} scaled to {correspond to a} 10~dB signal-to-noise-ratio (SNR).
We report the results when training with both REINFORCE (used in previous work) and our DQN optimisation approach (Section~\ref{sec:optim}).
All models are evaluated in both the training and evaluation environments {(Section~\ref{sec:acoustic_channel})}.
We report the accuracy per attribute: the average number of attributes correctly communicated out of all attribute combinations.
We also report the structural similarity of the emergent communication in terms of Spearman $\rho$ correlation between the input and message space, a metric known as topographic similarity or \textit{topsim} \citep{lazaridou-etal-2018,brighton-kirby-2006-understanding}.
This metric gives an indication of how compositional a learnt communication protocol is by finding the correlation between which message symbols are reused for the same input combinations.
For example, if we had to communicate the attributes ``red square'' and ``red triangle'' and used the sequence ``aaa'' to do so in both cases, we would get a high topsim. If, however, we used ``aaa'' in the first case and ``eee'' in the second to indicate the colour, we would get a low topsim.
For this metric, higher is better and the maximum value is 1.

\textbf{Experimental results.}
The results are presented in Table~\ref{tab:models}.
All the models optimised with DQN outperform their REINFORCE counterparts, {with particularly large improvements}  in compositionality (as measured by the topsim metric).
When looking only at the DQN results, we see {that} the acoustic models (rows 2 to 5) are consistently able to adapt to the noisy environment and outperform the \discrete{} model (row 1).
The end-to-end models {(with and without pre-training, rows 2 and 3)}
outperform the models using the phone recogniser (rows 4 and 5). This is to be expected as the Listener can directly adapt to the noisy environment by updating the weights of the ``hearing'' portion of the model -- this is not possible when using a fixed phone recogniser.
Nevertheless, by comparing the system in row 1 with those in rows 4 and 5, we see that the acoustic models that use the same phone recogniser as the \discrete{} model are able to adapt their communication protocol in order to mitigate the effects of the channel noise.
While the DQN models are not as reliant on discrete pre-training as the REINFORCE models, we still find it to ease the learning process and improve results -- illustrated by the relative performance difference between rows 2 and 3 for REINFORCE vs DQN.
The best-performing model overall is the \acousticee{} model, where the Speaker is pre-trained {(row 3)}.
While pre-trained models already have an established communication protocol, the models without pre-training have to learn to deal with the communication loss at the same time, increasing the difficulty of learning.
This is much more apparent in the REINFORCE models.
In terms of compositionality, the acoustic DQN models tend to have slightly higher topographic similarity than the \discrete{} model, reaching a similar performance to that achieved in the discrete-only study of \citep{auersperger-pecina-2022-defending}.

\begin{table*}[!tb]
    \caption{Accuracy of the discrete and acoustic agents evaluated in various acoustic environments. \acousticpr{} refers to the pre-trained acoustic + phone recogniser model. \textit{Meeting} and \textit{stairway} refer to the two evaluation rooms, where the \textit{stairway} has 
    {more}
    echos than the \textit{meeting} room. Each model was trained five 
    times. We observe a maximum standard deviation of 0.03 over all runs for all results.}
    \centering
    \small
    \begin{tabular}{l c c c c c c c c}
        \toprule
        \multicolumn{3}{c}{} & \multicolumn{3}{c}{No background noise} & \multicolumn{3}{c}{10~dB SNR background noise}  \\
        \cmidrule(lr){4-6} \cmidrule(lr){7-9}
        \textit{Model} & $L$ & \textit{no room} & \textit{training rooms} & \textit{meeting} & \textit{stairway} & \textit{training rooms} & \textit{meeting} & \textit{stairway} \\
        \midrule
        \discrete{}&	5&	0.992&	0.780&	0.794&	0.594&	0.651&	0.703&	0.604\\
        \acousticpr{} &	5&	0.945&	0.808&	0.798&	0.634&	0.726&	0.751&	0.664\\
        \midrule
        \discrete{}&	8&	0.998&	0.728&	0.801&	0.546&	0.564&	0.666&	0.544\\
        \acousticpr{} &	8&  0.950&	0.806&	0.826&	0.635& 0.677&	0.731&	0.640\\
        \bottomrule
    \end{tabular}
    \label{tab:noise}
\end{table*}

In the experiments that follow, we will focus on a subset of the different the DQN optimised models from Table~\ref{tab:models}.
Specifically, we will compare discrete model (row 1) to both the pre-trained \acousticee{} (row 3) and \acousticpr{} (row 5) approaches.
We choose the \acousticpr{} models as a direct comparison to the discrete models, as both are restricted by the performance of the static phone recogniser.
We also chose the best performing model overall (\acousticee{}) to see how the learnt communication protocol varies based on the Listener dynamics.

\subsection{Increasing noise and sequence length}

To better understand the difference in emergent language between discrete and acoustic communication, we evaluate the {\discrete} and {\acousticpr} models with various configurations of the lossy communication channel.
While \acousticpr{} does not perform as well as \acousticee{}, we choose to focus on \acousticpr{} in this set of experiments as it allows for a fairer comparison to the \discrete{} model.
Both these models are restricted by the imperfect phone recogniser.
The \acousticpr{} model must adapt without updating the weights of the listening component.
Therefore, it must develop a communication strategy to compensate for transmission errors over the noisy channel and static phone recogniser. These properties allow us to better see the differences in the communication protocols resulting from discrete vs acoustic training.

\textbf{Experimental setup.} Concretely, we consider performance in settings with a lossless communication channel (\textit{no room}), a set of rooms with no background noise, and the same set of rooms with 10~dB SNR background noise sampled from Clotho.
Three room setups are considered: an evaluation of the rooms seen during training, an evaluation of the unseen \textit{meeting} room, and an evaluation of the unseen \textit{stairway}.
The models are trained and evaluated with a maximum phone length $L$ of both $5$ and $8$ in order to see the effect of varying sequence lengths.

\begin{figure}[!b]
\centering
\vspace{-0.8cm}
\includegraphics[width=1\linewidth]{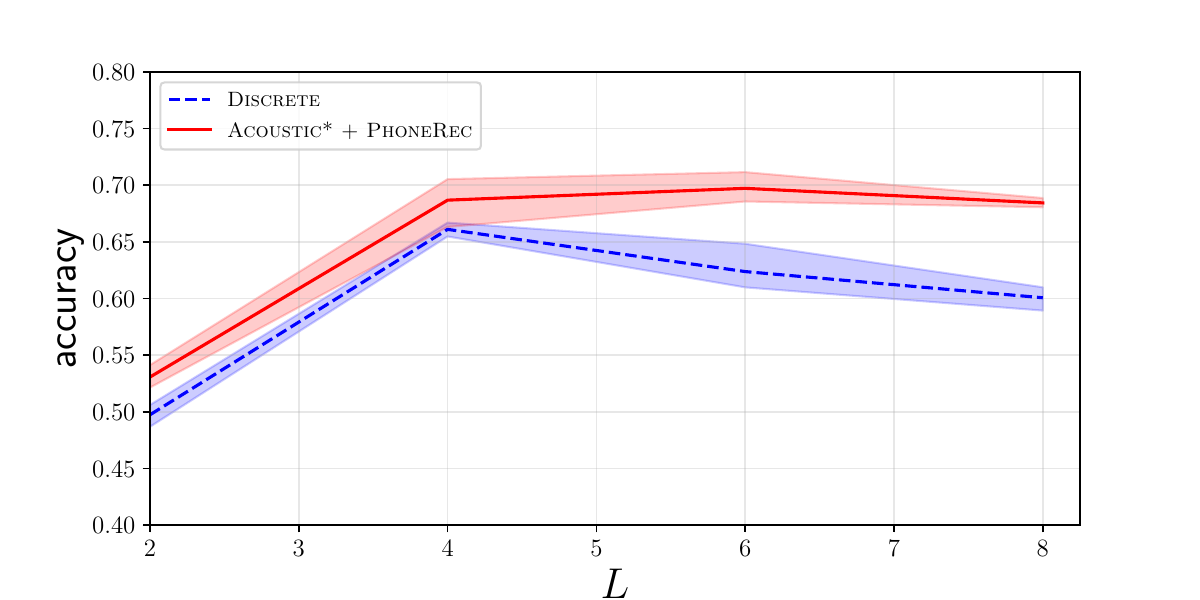}
\caption{Per-attribute accuracy as a function of maximum phone length $L$. The 95\% confidence bounds over five runs are shown.}
\label{fig:seq_len}
\end{figure}

\textbf{Experimental results.}
The results are shown in Table~\ref{tab:noise}.
We see that for a maximum phone length $L=5$, the \acousticpr{} models perform marginally better than the \discrete{} models in all the cases (with and without noise), as also was the case in Table~\ref{tab:models}.
Somewhat surprisingly, when background noise is present and the \discrete{} model is allowed to use $L=8$ symbols instead of $L=5$, the performance drops.
This is likely due to an increased probability that one or more phones are lost over the communication channel in longer sequences.
Despite this, due to the increased channel capacity when $L=8$, the \acousticpr{} model is able to counteract the information loss and retain performance comparable to its $L = 5$ counterpart.
This causes the \acousticpr{} model to do significantly better for longer sequences, with 13.3\% better relative performance
over the \discrete{} case 
in the \textit{meeting} and \textit{stairway} rooms with background noise.
We see similar results in Figure~\ref{fig:seq_len}, where we plot evaluation accuracy as a function of $L$: again we observe the \discrete{} model drops in performance after $L=4$, while the \acousticpr{} plateaus without a performance drop.
The performance drop at $L=2$ is due to the channel capacity falling below the total number of attribute combinations ($|S| < |\mathcal{P}|^L$).
This means the agents are unable to express all possible input combinations given the limited channel capacity.

Both the discrete and acoustic models perform best in the \textit{meeting} evaluation room, where the acoustic dynamics are relatively clean.
The models perform worst in the \textit{stairway} evaluation room, where there are a lot of surfaces for sound to echo off of.
The models also experience a large performance decrease when significant background noise is present, as can be seen by comparing the `No background noise' columns to the `10~dB SNR background noise' columns.

\subsection{Emergent compositionality and redundancy}

\belowrulesep=0ex
\begin{table}[!tb]
    \scriptsize
    \fontsize{8}{10}
    \vspace{0.075cm}
    \caption{Sample of phone sequences produced by the \discrete{} Speaker for $L=8$.
        Each entry corresponds to a combination of four attributes: $\boldsymbol s_1$ and $\boldsymbol s_2$ varying, while $\boldsymbol s_3$ and $\boldsymbol s_4$ are fixed.}
    \centering
    \vspace{-0.25cm}
    \scriptsize
    \hspace*{-0.25cm}\begin{tabular}{c c c c c c c}
        && \multicolumn{5}{c}{$\boldsymbol s_1$} \\[0.3em]
        && \multicolumn{1}{c}{0} & \multicolumn{1}{c}{1} & \multicolumn{1}{c}{2} & \multicolumn{1}{c}{3} & \multicolumn{1}{c}{4}\\
        \cmidrule{3-7} \rule{0pt}{1.2EM}
        \parbox[t]{0mm}{\multirow{6}{*}{\rotatebox[origin=c]{90}{$\boldsymbol s_2$}}} 
        & \parbox[t]{2mm}{\multirow{1}{*}{\rotatebox[origin=c]{90}{0}}}
        & \multicolumn{1}{|c}{{\textipa{auieueuo}}} 
        & {\textipa{uiieueoo}} 
        & {\textipa{oiaeueou}} 
        & {\textipa{aiaeueou}} 
        & {\textipa{iiaeeoeo}} \\[0.4em]
        
        & \parbox[t]{2mm}{\multirow{1}{*}{\rotatebox[origin=c]{90}{1}}} 
        & \multicolumn{1}{|c}{{\textipa{aiiuoooi}}} 
        & {\textipa{uiiuoooi}} 
        & {\textipa{oiiuoooi}} 
        & {\textipa{aiiuoooi}} 
        & {\textipa{iiiuoooi}} \\[0.4em]
        
        & \parbox[t]{2mm}{\multirow{1}{*}{\rotatebox[origin=c]{90}{2}}} 
        & \multicolumn{1}{|c}{{\textipa{uuieueou}}} 
        & {\textipa{uuieoeoi}} 
        & {\textipa{ouieueou}} 
        & {\textipa{auieoeoi}} 
        & {\textipa{iuieueou}} \\[0.4em]
        
        & \parbox[t]{2mm}{\multirow{1}{*}{\rotatebox[origin=c]{90}{3}}} 
        & \multicolumn{1}{|c}{{\textipa{aiieueoe}}} 
        & {\textipa{eiieueoo}} 
        & {\textipa{oiieueoo}} 
        & {\textipa{aiieueoe}} 
        & {\textipa{iiieueoe}} \\[0.4em]
        
        & \parbox[t]{2mm}{\multirow{1}{*}{\rotatebox[origin=c]{90}{4}}} 
        & \multicolumn{1}{|c}{{\textipa{aaieeueo}}} 
        & {\textipa{eaieoeoi}} 
        & {\textipa{oaieeouo}} 
        & {\textipa{aaieeoui}} 
        & {\textipa{iaieeouo}} \\[0.4em]
    \end{tabular}
    \label{tab:samples_discrete}
    
\vspace{0.3cm}

    \caption{Sample of phone sequences produced by the \acousticpr{} Speaker for $L=8$.
         Each entry corresponds to a combination of four attributes: $\boldsymbol s_1$ and $\boldsymbol s_2$ varying, while $\boldsymbol s_3$ and $\boldsymbol s_4$ are fixed.
         {The bigram [{\textbf{oi}}] has been highlighted in bold.}}
    \centering
    \vspace{-0.25cm}
    \scriptsize
    \hspace*{-0.25cm}\begin{tabular}{c c c c c c c}
        && \multicolumn{5}{c}{$\boldsymbol s_1$} \\[0.3em]
        && \multicolumn{1}{c}{0} & \multicolumn{1}{c}{1} & \multicolumn{1}{c}{2} & \multicolumn{1}{c}{3} & \multicolumn{1}{c}{4}\\
        \cmidrule{3-7} \rule{0pt}{1.2EM}
        \parbox[t]{0mm}{\multirow{6}{*}{\rotatebox[origin=c]{90}{$\boldsymbol s_2$}}} 
        & \parbox[t]{2mm}{\multirow{1}{*}{\rotatebox[origin=c]{90}{0}}}
        & \multicolumn{1}{|c}{{{auaa\textbf{oioi}}}} 
        & {{eaao\textbf{oioi}}} 
        & {{oaa\textbf{oioi}o}} 
        & {{aea\textbf{oioi}o}} 
        & {{ioa\textbf{oioi}i}} \\[0.4em]
        
        & \parbox[t]{2mm}{\multirow{1}{*}{\rotatebox[origin=c]{90}{1}}} 
        & \multicolumn{1}{|c}{{{aoa\textbf{oioi}o}}} 
        & {{uoao\textbf{oioi}}} 
        & {{ooa\textbf{oioi}o}} 
        & {{aoa\textbf{oioi}o}} 
        & {{ioa\textbf{oioi}o}} \\[0.4em]
        
        & \parbox[t]{2mm}{\multirow{1}{*}{\rotatebox[origin=c]{90}{2}}} 
        & \multicolumn{1}{|c}{{{auaao\textbf{oi}o}}} 
        & {{uuaaoo\textbf{oi}}} 
        & {{uuaaoo\textbf{oi}}} 
        & {{auaoo\textbf{oi}i}} 
        & {{iuaao\textbf{oi}o}} \\[0.4em]
        
        & \parbox[t]{2mm}{\multirow{1}{*}{\rotatebox[origin=c]{90}{3}}} 
        & \multicolumn{1}{|c}{{{eeauio\textbf{oi}}}} 
        & {{eea\textbf{oioi}o}} 
        & {{oea\textbf{oioi}o}} 
        & {{aea\textbf{oioi}o}} 
        & {{iea\textbf{oioi}o}} \\[0.4em]
        
        & \parbox[t]{2mm}{\multirow{1}{*}{\rotatebox[origin=c]{90}{4}}} 
        & \multicolumn{1}{|c}{{{aaaao\textbf{oi}o}}} 
        & {{eaaoo\textbf{oi}i}} 
        & {{oaaoo\textbf{oi}i}} 
        & {{aaaoo\textbf{oi}i}} 
        & {{iaaao\textbf{oi}o}} \\[0.4em]
    \end{tabular}
    \label{tab:samples_acoustic_asr}

\vspace{0.3cm}

    \caption{{Sample of phone sequences produced by the \acousticee{} Speaker for $L=8$.
         Each entry corresponds to a combination of four attributes: $\boldsymbol s_1$ and $\boldsymbol s_2$ varying, while $\boldsymbol s_3$ and $\boldsymbol s_4$ are fixed.}
         {The bigram [{\textbf{oi}}] has been highlighted in bold.}}
    \centering
    \vspace{-0.25cm}
    \scriptsize
    \hspace*{-0.25cm}\begin{tabular}{c c c c c c c}
        && \multicolumn{5}{c}{$\boldsymbol s_1$} \\[0.3em]
        && \multicolumn{1}{c}{0} & \multicolumn{1}{c}{1} & \multicolumn{1}{c}{2} & \multicolumn{1}{c}{3} & \multicolumn{1}{c}{4}\\
        \cmidrule{3-7} \rule{0pt}{1.2EM}
        \parbox[t]{0mm}{\multirow{6}{*}{\rotatebox[origin=c]{90}{$\boldsymbol s_2$}}}   
        & \parbox[t]{2mm}{\multirow{1}{*}{\rotatebox[origin=c]{90}{0}}}
        & \multicolumn{1}{|c}{{{auaee\textbf{oi}o}}} 
        & {{eeaoe\textbf{oi}e}} 
        & {{oeae\textbf{oioi}}} 
        & {{aeae\textbf{oioi}}} 
        & {{ieae\textbf{oioi}}} \\[0.4em]
        
        & \parbox[t]{2mm}{\multirow{1}{*}{\rotatebox[origin=c]{90}{1}}} 
        & \multicolumn{1}{|c}{{{aiau\textbf{oioi}}}} 
        & {{eiao\textbf{oioi}}} 
        & {{\textbf{oi}au\textbf{oioi}}} 
        & {{aiao\textbf{oioi}}} 
        & {{iia\textbf{oioi}e}} \\[0.4em]
        
        & \parbox[t]{2mm}{\multirow{1}{*}{\rotatebox[origin=c]{90}{2}}} 
        & \multicolumn{1}{|c}{{{uuaeeoio}}} 
        & {{uuae\textbf{oioi}}} 
        & {{ouaee\textbf{oi}o}} 
        & {{auae\textbf{oioi}}} 
        & {{iuaee\textbf{oi}o}} \\[0.4em]
        
        & \parbox[t]{2mm}{\multirow{1}{*}{\rotatebox[origin=c]{90}{3}}} 
        & \multicolumn{1}{|c}{{{aeaee\textbf{oi}o}}} 
        & {{eeaoe\textbf{oi}e}} 
        & {{oeaee\textbf{oi}o}} 
        & {{aeae\textbf{oioi}}} 
        & {{ieaee\textbf{oi}o}} \\[0.4em]
        
        & \parbox[t]{2mm}{\multirow{1}{*}{\rotatebox[origin=c]{90}{4}}} 
        & \multicolumn{1}{|c}{{{aaaee\textbf{oi}o}}} 
        & {{eaaoe\textbf{oi}e}} 
        & {{oaaee\textbf{oi}o}} 
        & {{aaae\textbf{oioi}}} 
        & {{iaaee\textbf{oi}o}} \\[0.4em]
    \end{tabular}\vspace{-0.3cm}
    \label{tab:samples_acoustic_e2e}
\end{table}

The learnt communication protocols of each agent from the previous section are now investigated qualitatively. 
The goal here is to take a deeper dive into the emergent communication protocols of each model, in an attempt to understand how and why the acoustic models outperform the discrete ones.
To do this, we take samples of the phone sequences produced by the Speaker agent of each model.

{Table~\ref{tab:samples_acoustic_asr} and Table~\ref{tab:samples_discrete} respectively show samples of the learnt communication strategy of the \acousticpr{} and \discrete{} case with $L=8$.
For comparison, we also show samples of the \acousticee{} Speaker in Table~\ref{tab:samples_acoustic_e2e}.
{Each entry in the tables correspond to the phone sequence generated by the Speaker for different attribute values.
$s_1$ and $s_2$ are varied from 0 to 5 across the columns and rows, respectively, while $s_3$ and $s_4$ are fixed.
Following the example from Section~\ref{sec:environment} where $s_1$ represents a colour, each column could represent a colour (e.g. \textit{red}, \textit{green}, \textit{blue}).}

{By comparing the phones used per utterance in Table~\ref{tab:samples_discrete} to Tables~\ref{tab:samples_acoustic_asr}
and \ref{tab:samples_acoustic_e2e}, it} 
is immediately clear that both acoustic Speakers tend to use fewer unique phones per utterance, and also tend to repeat phonetic unigrams, bigrams and trigrams.
A trigram is a contiguous sequence of three units, a bigram a sequence of two units, and a unigram a single unit.
{For instance, as shown in Table~\ref{tab:samples_acoustic_asr}, to communicate this specific combination of $s_3$ and $s_4$, the Speaker tends to use the repeated bigram [\textipa{{oi}}].}
The same behaviour is observed in Table~\ref{tab:samples_acoustic_e2e}, with both acoustic Speakers repeating this bigram, despite being trained independently and with a completely different Listener setup.
This is interesting, as the two acoustic models have no direct motivation to learn similar protocols, other than overcoming the transmission errors of the noisy communication channel.}

The average number of repeated phones in the \textsc{Acoustic*~+~Pho-neRec}
{Speaker's} utterances is 2.89, while the \discrete{} Speaker has 3.19 repeated phones per utterance.
This is an indication that the acoustic Speaker is learning a redundant communication protocol, assisting the coherency of the Listener through repetition.
Table~\ref{tab:repeats} shows the number of repeated bigrams and trigrams for each model.
Both acoustic models tend to repeat bigrams and trigrams, with the \acousticee{} model having fewer repeats.
This is likely due to repetition not being as necessary when the network weights of the ``hearing'' portion may be updated.
This repetition is not as present in the discrete case, where the communication strategy uses more unique phones per utterance: on average, the \discrete{} Speaker uses 4.074 unique characters per utterance, with the \acousticpr{} Speaker using fewer phones at 3.649.

All the acoustic and discrete Speakers exhibit high levels of compositionality.
For example, all three cases {in Tables~\ref{tab:samples_discrete} to \ref{tab:samples_acoustic_e2e}} begin with [a] where $\ s_1 = 3$ and [i] where $\ s_1 = 4$.
Another example can be seen in that the second phone tends to correspond with $s_2$ in all cases.
These tables qualitatively show the levels of topographic similarity measured in Section~\ref{sec:approaches} {(specifically Table~\ref{tab:models})}.

\begin{table}[!t]
    \caption{Number of repeated bigrams and trigrams per utterance for each model ($L=8$).}
    \centering
    \small
    \begin{tabular}{l c c}
        \toprule
        \textit{Model} & \textit{bigrams} & \textit{trigrams} \\
        \midrule
        \discrete{} & 1.623 & 0.277 \\
        \acousticpr{} & 2.680 & 0.935 \\
        \acousticee{} & 2.073 & 0.554 \\
        \bottomrule
    \end{tabular}
    \label{tab:repeats}
\end{table}

%% file: s5_discussion.tex
\section{SUMMARY AND CONCLUSION}
\label{sec:discussion}

This paper has laid the foundation for investigating whether we can observe emergent language between agents using a continuous acoustic communication channel trained through RL.
Our acoustic environment allows us to see how the channel conditions affect the language that emerges.
In this environment, we observe that the acoustic Speaker learns redundancy which improves Listener coherency.
We also observe how the Speaker learns to update the emergent communication protocol to minimise errors due to the imperfect Listener.
We proposed two variations of the acoustic setup, one with a static phone recogniser and an end-to-end variation.
When analysing the emergent communication protocols, we found that agents trained with a communication channel learn similar strategies of repetition, often repeating the same bigrams.
Alongside this, the acoustic agents learn to be slightly more compositional than the purely discrete agents.
These are examples of emergent linguistic behaviour that is not modelled in a purely discrete setting.

This work expands on \citep{gao-etal-2020-spoken, komatsu-2022-automatic} (see Section~\ref{sec:introduction}), where the communication is restricted to fixed audio snippets from a discovered dictionary.
In contrast, our Speaker agent has the ability to generate unique audio waveforms.
On the other hand, our Speaker can only generate sequences based on a fixed phone set (which is then passed over a continuous acoustic channel).
This differs from earlier work~\citep{howard-etal-2014-learning,asada-2016-modeling, rasilo-rasanen-2017-online} that considered a Speaker that learns a full articulation model 
in an effort to come as close as possible to imitating an utterance from a caregiver; this allows a Speaker to generate arbitrary learnt units.
We have thus gone further than \citep{gao-etal-2020-spoken,komatsu-2022-automatic} but not as far as these older studies.
Despite these shortcomings, our approach has the benefit in that it is formulated in a modern multi-agent RL setting and can be easily extended.
Future work can therefore consider whether articulation can be learnt as part of our model 
-- possibly using imitation learning to guide the agent's exploration of the very large action space of articulatory movements.
This speaks to the modularity of our setup, where any number of components may be switched out.

We have also shown that DQN is better suited for this sort of referential game environment than REINFORCE, reaching similar performance to {that achieved in the discrete-only study of} \citep{auersperger-pecina-2022-defending}.
We highly recommend
other practitioners working on emergent communication to consider DQN as an alternative method for optimisation -- perhaps in combination with the models used in \citep{auersperger-pecina-2022-defending}.

In the experiments carried out in this study, we only considered a single communication round between two agents.
This could be expanded to a setup where each agent has both a speaking and listening module, and these composed agents then communicate with one another.
Future work could therefore consider multi-round communication games between {two} or more agents.
Such games would extend our work to the full MARL problem, where agents would need to ``speak'' to and ``hear'' each other to solve a common task.